\documentclass[pdflatex,sn-nature]{sn-jnl}

\usepackage{graphicx}%
\usepackage{multirow}%
\usepackage{amsmath,amssymb,amsfonts}%
\usepackage{amsthm}%
\usepackage{mathrsfs}%
\usepackage[title]{appendix}%
\usepackage{xcolor}%
\usepackage{textcomp}%
\usepackage{manyfoot}%
\usepackage{booktabs}%
\usepackage{algorithm}%
\usepackage{algorithmicx}%
\usepackage{algpseudocode}%
\usepackage{listings}%
\usepackage{fix-cm}

\begin{document}

\title[Article Title]{Digital Gene: Learning about the Physical World through Analytic Concepts}


\author{\fnm{Jianhua} \sur{Sun}}\email{gothic@sjtu.edu.cn}
\author{\fnm{Cewu} \sur{Lu}}\email{lucewu@sjtu.edu.cn}

\affil{\orgname{Shanghai Jiao Tong University, Shanghai Innovation Institute}}







\abstract{Reviewing the progress in artificial intelligence over the past decade, various significant advances (\textit{e.g.} object detection, image generation, large language models) have enabled AI systems to produce more semantically meaningful outputs and achieve widespread adoption in internet scenarios. Nevertheless, AI systems still struggle when it comes to understanding and interacting with the physical world. This reveals an important issue: relying solely on semantic-level concepts learned from internet data (\textit{e.g.} texts, images) to understand the physical world is far from sufficient -- machine intelligence currently lacks an effective way to learn about the physical world. This research introduces the idea of \textbf{analytic concept} -- representing the concepts related to the physical world through programs of mathematical procedures, providing machine intelligence a portal to perceive, reason about, and interact with the physical world. Except for detailing the design philosophy and providing guidelines for the application of analytic concepts, this research also introduce about the infrastructure that has been built around analytic concepts. I aim for my research to contribute to addressing these questions: What is a proper abstraction of general concepts in the physical world for machine intelligence? How to systematically integrate structured priors with neural networks to constrain AI systems to comply with physical laws?}

\maketitle

\section{Background: AI in Physical World Applications}

\begin{figure}[!htp]
  \centering
  \includegraphics[width=\linewidth, trim={0 4cm 0 4cm},clip]{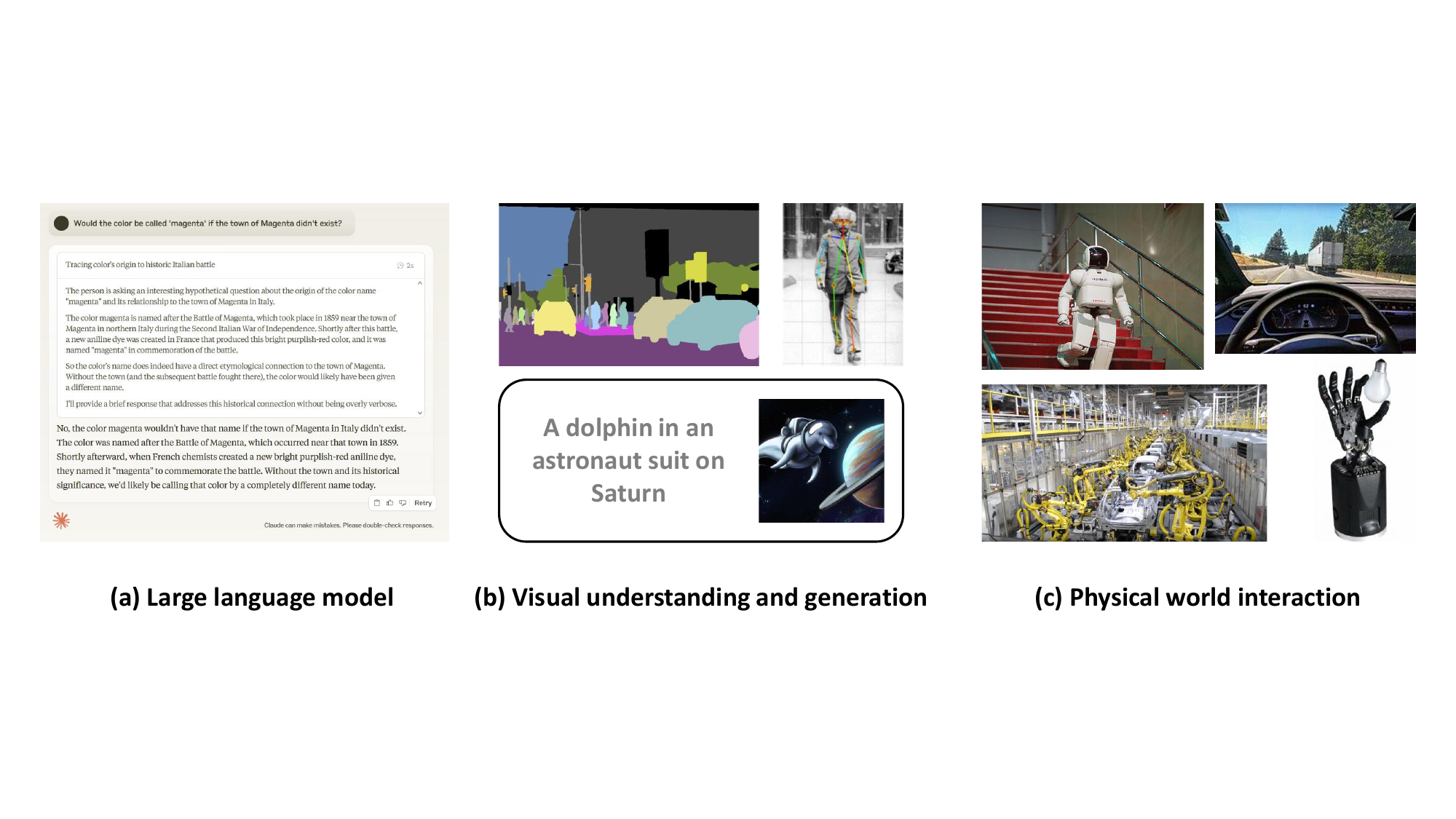}
  \caption{Examples of applications with high (a,b) and low (c) failure tolerance. The former typically involves tasks related to semantic understanding and generation, while the latter primarily includes tasks involving interaction with the physical world.}
  \label{fig:taskcat}
\end{figure}

Human intelligence are truly impressive. Its versatility and adaptability empower individuals to excel across a vast array of matters, spanning from the routine challenges of daily life and the complexities of production to the imaginative processes of artistic creation and the analytical rigor of scientific research. Considering the outcomes, tasks involving human participation can be categorized into those with high failure tolerance (failures do not result in economic losses) and those with low failure tolerance (failures result in economic losses). Fig.~\ref{fig:taskcat} provides several examples.

While humans excel at handling all types of tasks, a review of the past decade’s advancements in artificial intelligence reveals that despite that AI systems has already played a dominant role in many of the first category of tasks, its effectiveness is still limited in applications with low failure tolerance, most of which involve interaction with the physical world. This fact reveals that although current AI can assist humans for efficiency, it cannot fully replace humans in completing intricate tasks in the physical world.

Then, what do highly autonomous systems for applications in the physical world look like today? Like industrial production lines for manufacturing or high-speed trains and automated guided vehicles (AGVs) used in transportation, these autonomous systems rely on pre-programmed instructions or procedures to perform fixed tasks in structured environments. In this way, such systems can safely and reliably replace humans in applications in the physical world.

Given these aforementioned findings, two compelling questions emerge: What do we aim to achieve by adopting AI in physical world applications compared to traditional automation systems? And why are current AI systems not yet well prepared for physical world applications? I suggest the answers lie in the versatility and controllability of a system.

On one hand, the data-driven paradigm of AI offers better generalization compared to controlling with programmatic rules, inspiring the community to look to AI for creating more versatile systems that can quickly adapt to diverse tasks. However, it has been observed that the impressive generalization capabilities of current AI systems (such as LLMs) remain confined to the semantic level and fall short when it comes to accurately understanding and interacting with the physical world. This undermines the inherent advantages of AI approaches.

On the other hand, the data-driven paradigm of AI has a inherent disadvantage: it is difficult to control an AI to produce outputs strictly comply with rules given by users. This instability significantly hinders the application of AI in the physical world, where even a 1\% failure rate can result in substantial economic losses in some cases like industrial production and autonomous driving.

Therefore, to better adapt AI systems for physical world applications, I aim for my research to these questions: How to enhance machine intelligence to generally perceive, reason about, and interact with the physical world? How to systematically constrain AI systems to comply with physical laws?

\section{Motivation and Significance of Analytic Concepts}

Before delving into how analytical concepts are designed and utilized to achieve more generalized and controllable AI to perceive and interact with the physical world, let us first explore why current multi-modal large language models (MLLMs), which exhibit great generalization capability in visual understanding and generation, fall short in dealing with tasks in the physical world. 

\begin{figure}[!htp]
  \centering
  \includegraphics[width=\linewidth, trim={0 5cm 0 4cm},clip]{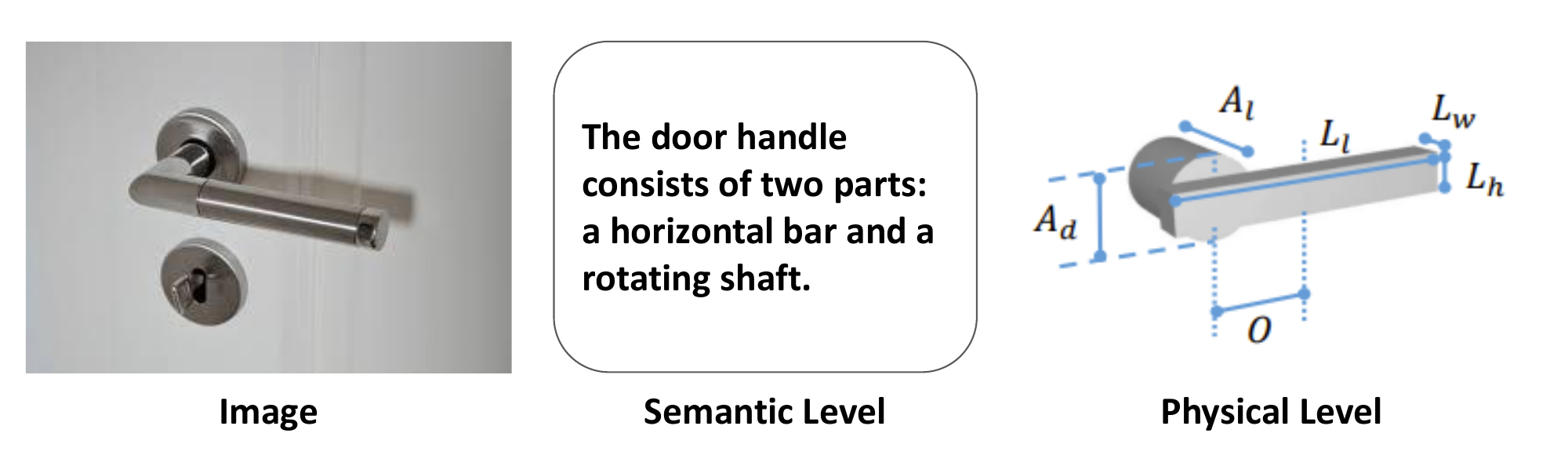}
  \caption{Concepts represented at semantic level \textit{vs.} physical level.}
  \label{fig:semantics-physics}
\end{figure}

A critical point here is that current MLLMs learn at the semantic level, without aligning with the physical world. They often receive images and texts as input, learn semantic-level embeddings which are grounded on natural language, and are supervised by internet tasks (reconstruction, generation, VQA). Such training paradigm allows AI to thoroughly capture the semantic-level concepts embedded within image and text data, but these concepts are not aligned with the physical quantities in the real world. A straightforward example is in Fig.~\ref{fig:semantics-physics}: an MLLM is able to figure out \textit{what the handle looks like} from the image but cannot precisely specify \textit{its geometric parameters in meters}. 

To enable MLLMs to possess similar generalization capability at the physical level as that at the semantic level, an intuitive approach is to directly use data collected in the physical world for training. However, this is not feasible in practice, as the cost is prohibitively high to collect data in the physical world and the amount of data required far exceeds that needed for training large language models. Besides, this purely data-driven approach makes it difficult to achieve strict control over the output of AI systems.

\begin{figure}[!htp]
  \centering
  \includegraphics[width=\linewidth, trim={0 5cm 0 5cm},clip]{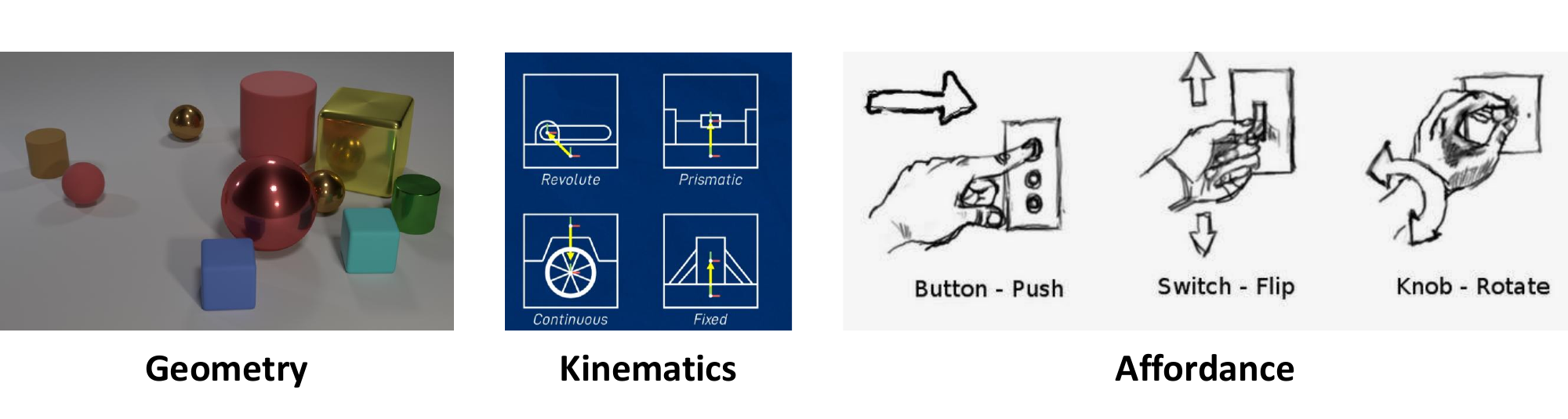}
  \caption{Common categories of physical world concepts.}
  \label{fig:conceptkinds}
\end{figure}

Therefore, I try to seek a clever way to enable AI to learn about the physical world effectively, namely \textbf{Analytic Concepts}. Analytic concepts are explicit representations of physical world concepts, expressed through programs of mathematical procedures. These physical world concepts include geometry, kinematics, affordance, \textit{etc.}, see Fig.~\ref{fig:conceptkinds}. The idea of analytic concept originates from the simulation of the physical world by computers, which uses mathematical models composed of equations or procedures that duplicate the concepts within the physical world. Given this analytic form of representation, leveraging analytic concepts in guiding AI to learn about the physical world offers the following key benefits:

\begin{itemize}
    \item \textbf{Analytic concepts provide neural networks with physical models of physical world concepts}, firmly aligning the learned features and inference results of neural networks with physical quantities. Neural networks can treat physical world understanding tasks as inferring programs of analytic concepts which form an analog of the given physical scene. Then machine intelligence can execute rigorous mathematical reasoning on these programs to derive desired results. This paradigm will concretely ensure the behavior of AI comply with physical laws and rules within the analytic concepts. 
    
    \item \textbf{Analytic concepts provide neural networks with a high-level abstraction of physical world concepts}, enabling AI to adapt more flexibly to a wide variety of task scenarios in the physical world. Revisiting the success of large language models, a significant factor behind their strong generalization capabilities in semantic understanding tasks is that natural language serves as a high-level abstraction, enabling words to generate infinite meanings through a finite set of syntax. Similarly, analytic concepts, serving as programmatic representations of essential physical concepts, can be composed under program syntax to cover an infinite variety of physical world scenarios. This property of analytic concepts equips AI with strong generalization capabilities in understanding the physical world. Besides, analytic concepts are effectively aligned with other modalities, such as language and images, further enhancing the AI's ability to generalize across both physical and semantic concepts.
    
    \item \textbf{Analytic concepts provide neural networks with structured priors of commonsense knowledge}, improving AI to reason about and interact with the physical world in terms of few-shot learning, interpretability and controllability. The physical world is structured and compositional, as scenes are made of objects, objects are made of parts, and knowledge like relations, attributes and functionalities are expressed through objects/parts. These structured priors can be seamlessly integrated in the programs of analytic concepts, which are subsequently combined with neural networks for physical world understanding. With the aid of these structured priors, AI systems are able to better handle novel objects/scenes and perform in ways that adheres to human commonsense cognition.
\end{itemize}

\begin{figure}[!htp]
  \centering
  \includegraphics[width=\linewidth, trim={0 0cm 0 0cm},clip]{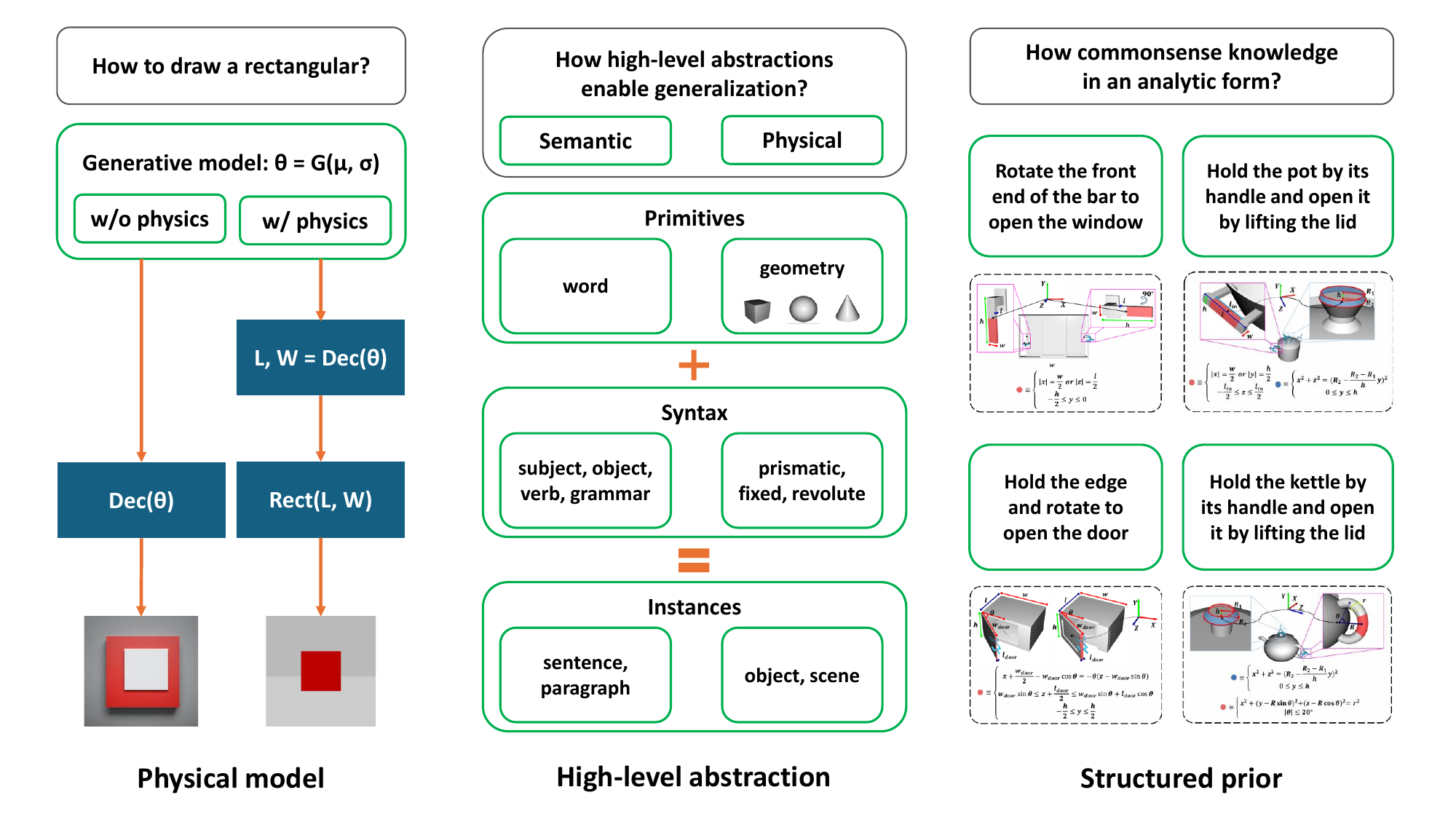}
  \caption{Illustrations of the significance of analytic concept's benefits.}
  \label{fig:conceptbenefit}
\end{figure}

Fig.~\ref{fig:conceptbenefit} provides several straightforward illustrations that demonstrate the significance of these benefits.
In summary, by integrating analytic concepts into AI systems, we seek to unlock substantial potential of AI in applications that demand a deep and precise understanding of the physical world, and enhance the AI's generalizability and controllability in the physical world. 

\textit{Throughout the history of scientific development, expressing abstract concepts in an analytic form has held profound significance. For example, ancient people abstracted the geometric concepts from natural objects like \textit{circle} and \textit{rectangular}, and Euclid proposed an axiomatic system, namely Euclidean geometry, introducing a well-defined way to formalize geometric concepts. While Euclidean geometry offered a systematic approach to the study of geometry, it was not until the advent of analytic geometry that scientists, for the first time, closely coupled the study of geometry with algebra, endowing geometric concepts with quantifiable and computable properties. This greatly broadened the scope and depth of the research of geometry, laying the foundation for fields such as calculus and computer graphics, while also offering critical mathematical tools for disciplines like physics. Similarly, the vision of this research lies in extending AI systems from learning semantic concepts to analytic concepts, thereby advancing machine intelligence's understanding of the physical world to a deeper level and finally empowering research and applications in the physical world such as robotics, autonomous driving and physical simulation.}

\section{Analytic Concepts: Philosophy and Techniques}

In this section, we provide a detailed exposition of the methodology for developing analytic concepts: What are the design philosophy of analytic concepts? How to develop techniques that model the physical world through analytical concepts and employ these concepts to enable physical-world interactions? How should the idea of analytic concepts be applied to a wide range of tasks?

\subsection{Philosophy: Analytically Represent General Physical Concepts in Daily Objects}

We begin our study of analytical concepts from the perspective of object geometry. The underlying motivation is straightforward: objects serve as the fundamental units for machine intelligence's perception and interaction with the physical world, and geometry - emerging from the base physical quantity of \textsc{Length} - measures the spatial dimensions in the physical world while the remaining mechanical quantities can be formulated based on spatial representations.

\begin{wrapfigure}{r}{0.35\textwidth}
\vspace{-20pt}
\includegraphics[width=\linewidth, trim={12cm 6cm 12cm 5cm},clip]{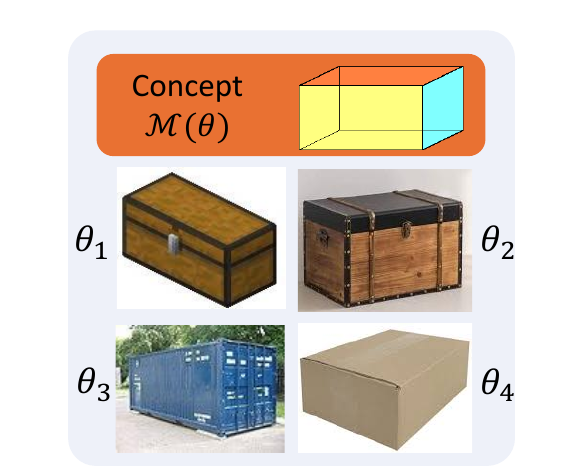} 
\caption{The concept \textit{cuboid} expresses the shared commonalities of the four different boxes and reflect how such commonalities vary across them through parameter $\theta=(l,w,h)$.}
\label{fig:conceptexample}
\end{wrapfigure}

Concepts, as abstractions of numerous similar instances, encapsulate their common characteristics while should also be capable of reflecting how such commonalities vary across distinct instances (see Fig.~\ref{fig:conceptexample}). A proper representation of concepts for machine intelligence ought to manifest this nature and need to be analytic - expressed through programs of mathematical procedures - to enable computational tractability in machine processing. 

\begin{figure}[!htp]
  \centering
  \includegraphics[width=\linewidth, trim={0 7cm 0 0cm},clip]{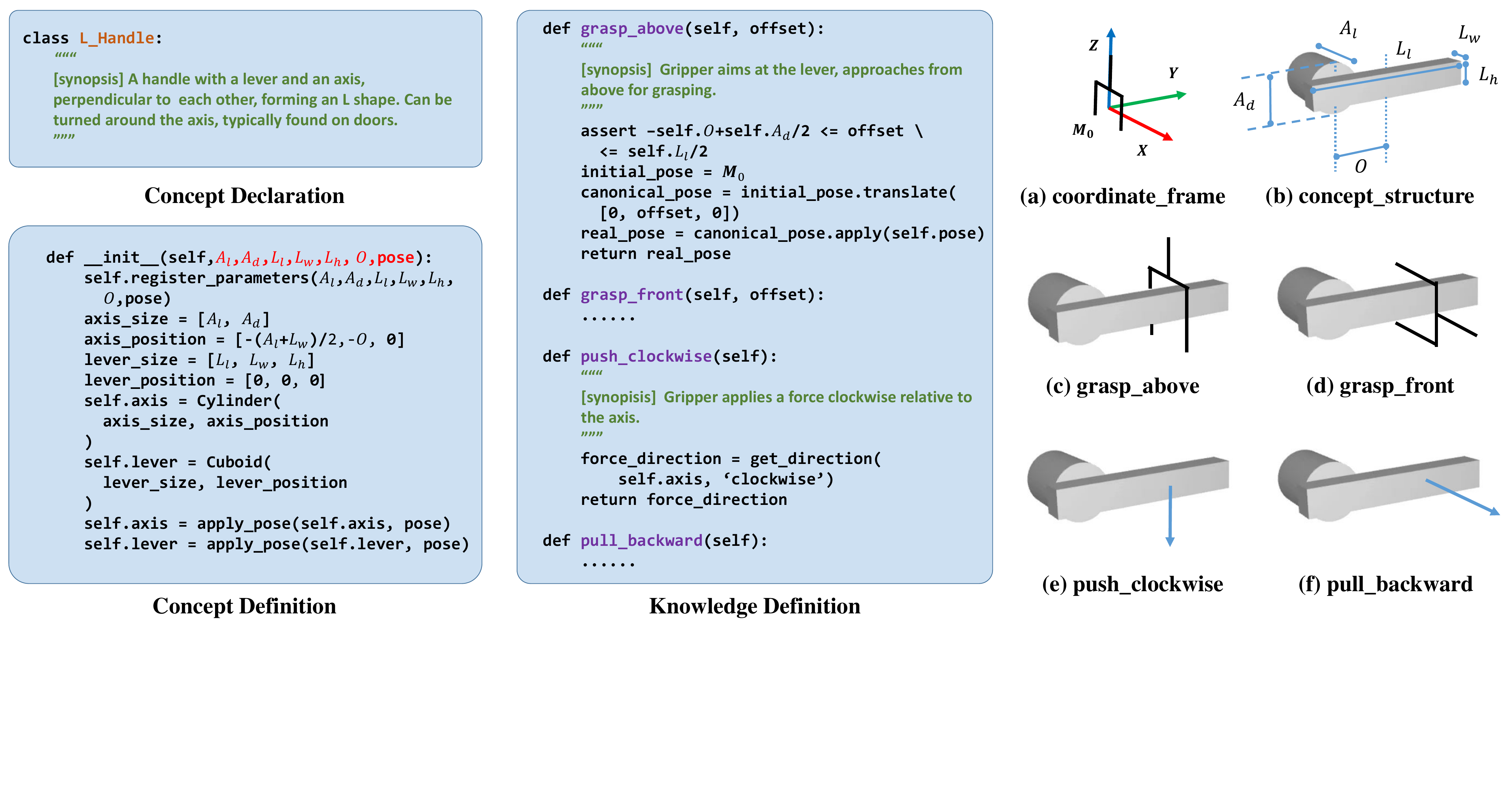}
  \caption{Implementation of analytic concept in Python \texttt{Class}. \texttt{L\_Handle} as example.}
  \label{fig:conceptdefinition}
\end{figure}

\texttt{Class}, which are parameterized through templates, offers an ideal formulation for expressing analytical concepts. The member attributes and functions in \texttt{Class} express the essential commonality among different instances belonging to the same class, and the \texttt{Class} parameters - variables in member attributes and functions - express the variances of different instances belonging to the same class. Meanwhile, \texttt{Class} is supported in mainstream programming languages. By implementing a concept as a \texttt{Class} like showing in Fig.~\ref{fig:conceptdefinition}, we establish a formal framework for abstraction of physical concepts in machine intelligence systems.

\begin{figure}[!htp]
  \centering
  \resizebox{\linewidth}{!}{\includegraphics[width=\linewidth, trim={0 0cm 0 0cm}, clip]{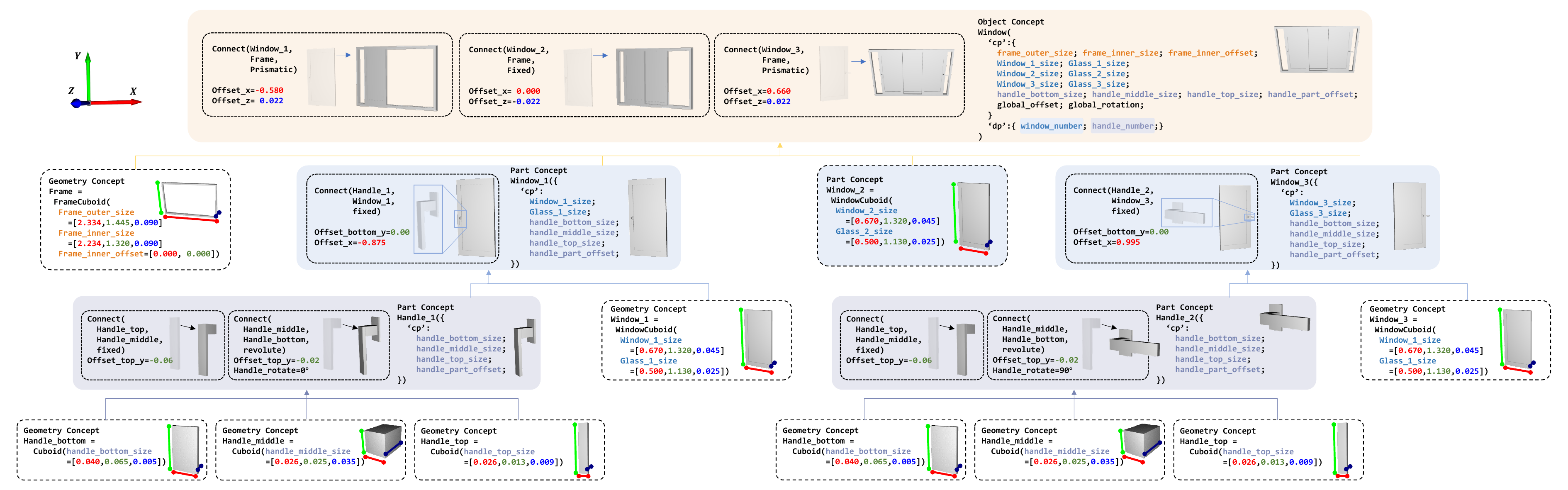}}
  \caption{Hierarchical definition of analytic concepts. White denotes basic geometric concepts, purple and blue denote part-level concepts at different scale, and yellow denotes object-level concepts. Zoom in for a clear view.}
  \label{fig:hierachydesign}
\end{figure}

This templated representation of analytic concepts offers significant design advantages. By leveraging the principles of encapsulation, inheritance, and polymorphism, it enables hierarchical definition from basic geometric concepts like \textit{cuboids} and \textit{cylinders} all the way up to part-level concepts like \textit{handles} and \textit{frames} and object-level concepts like \textit{windows} and \textit{doors}, as shown in Fig.~\ref{fig:hierachydesign}. That is, defining a new concept involves the invocation and reuse of existing concepts, forming a knowledge graph-like structure that facilitates efficient indexing while minimizing redundant definition efforts. 

Beyond expressing concepts of geometry in an analytic form, we naturally aim to describe other physical world concepts such as axial rotation, graspness, and \textit{etc.} Given that these concepts stem from geometric ones, we treat them as structured priors or knowledge of geometric concepts, implemented as member functions within the concept template, see Fig.~\ref{fig:conceptdefinition}-[Knowledge Definition].

In this section, we primarily focus the discussion on object-level concepts, as they constitute the most fundamental parts for perception and interaction between machine intelligences (e.g., robots) and the physical world. 

\subsection{Learning with Analytic Concepts}

\begin{figure}[!htp]
  \centering
  \resizebox{\linewidth}{!}{\includegraphics[width=\linewidth, trim={0 2cm 0 1cm}, clip]{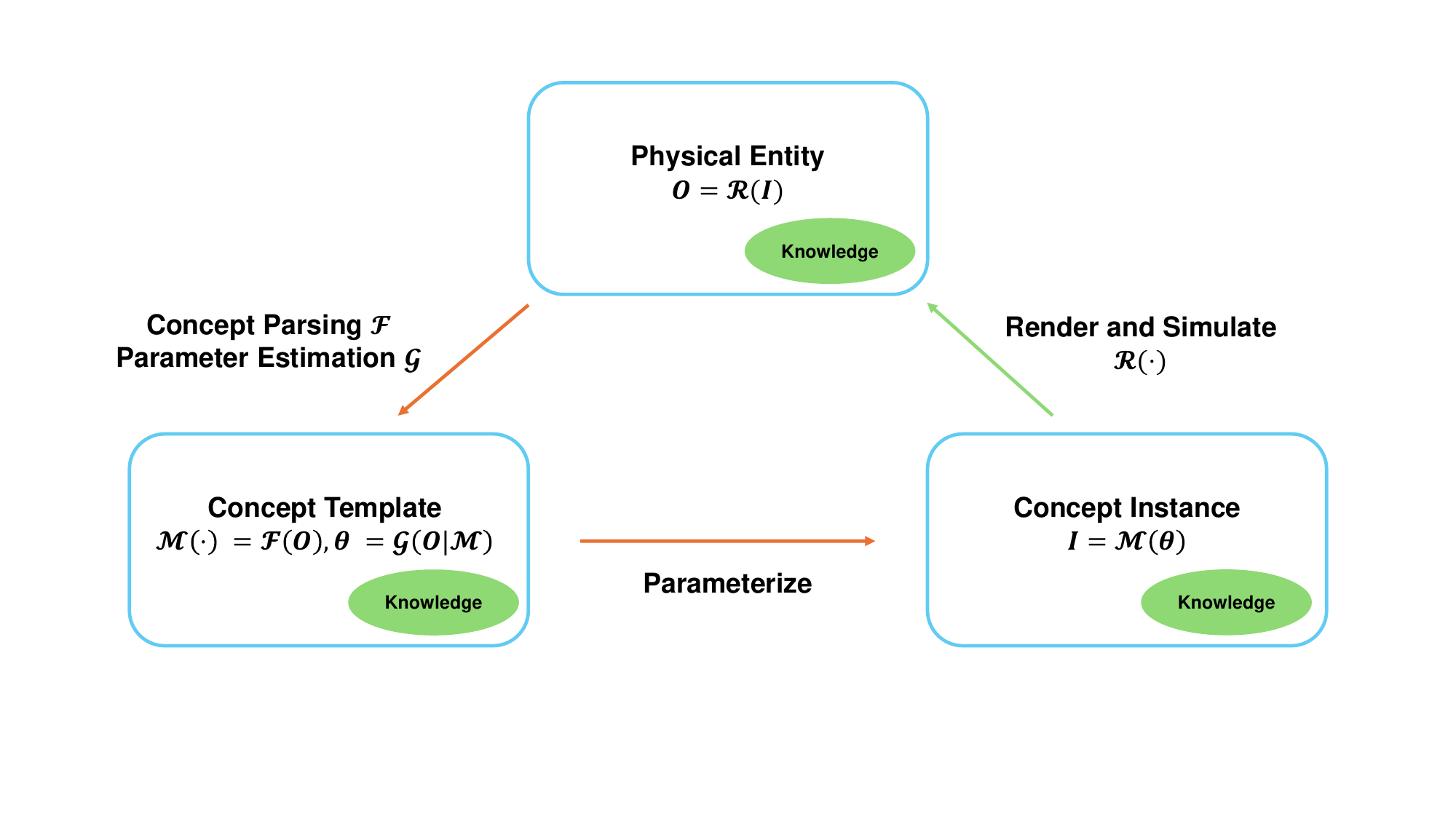}}
  \caption{The bidirectional alignment between physical entities and analytic concepts. The orange arrows demonstrate a forward process that abstracting a physical entity as a concept instance, and the green arrow demonstrates an inverse process that producing a physical entity according to a concept instance. Knowledge within concept templates defined trough template parameters is propagated as structured priors that align with concept instances and physical entities.}
  \label{fig:concept2physics}
\end{figure}

By expressing analytic concepts as \texttt{Class} templates, machine intelligence now possesses a portal to achieve perception, reasoning, and interaction with the physical world. This benefits from the most essential property of analytical concepts — their ability, as physical models of real-world concepts, to achieve bidirectional alignment with physical entities. Fig.~\ref{fig:concept2physics} provides a schematic of how physical entities are abstracted as concept instances and how we use concept instances to produce physical entities. In the remaining parts of this section, we first explain the forward and inverse process of learning with analytic concepts in detail, and then discuss about how both forward and inverse process can be integrated together to dynamically model complex physical worlds. Finally, we introduce ways to prepare training data for concept learning.

\subsubsection{From Physics to Concepts (Forward): Grounding, Reasoning and Creating}

\begin{figure}[!htp]
  \centering
  \resizebox{\linewidth}{!}{\includegraphics[width=\linewidth, trim={0 3cm 0 2cm}, clip]{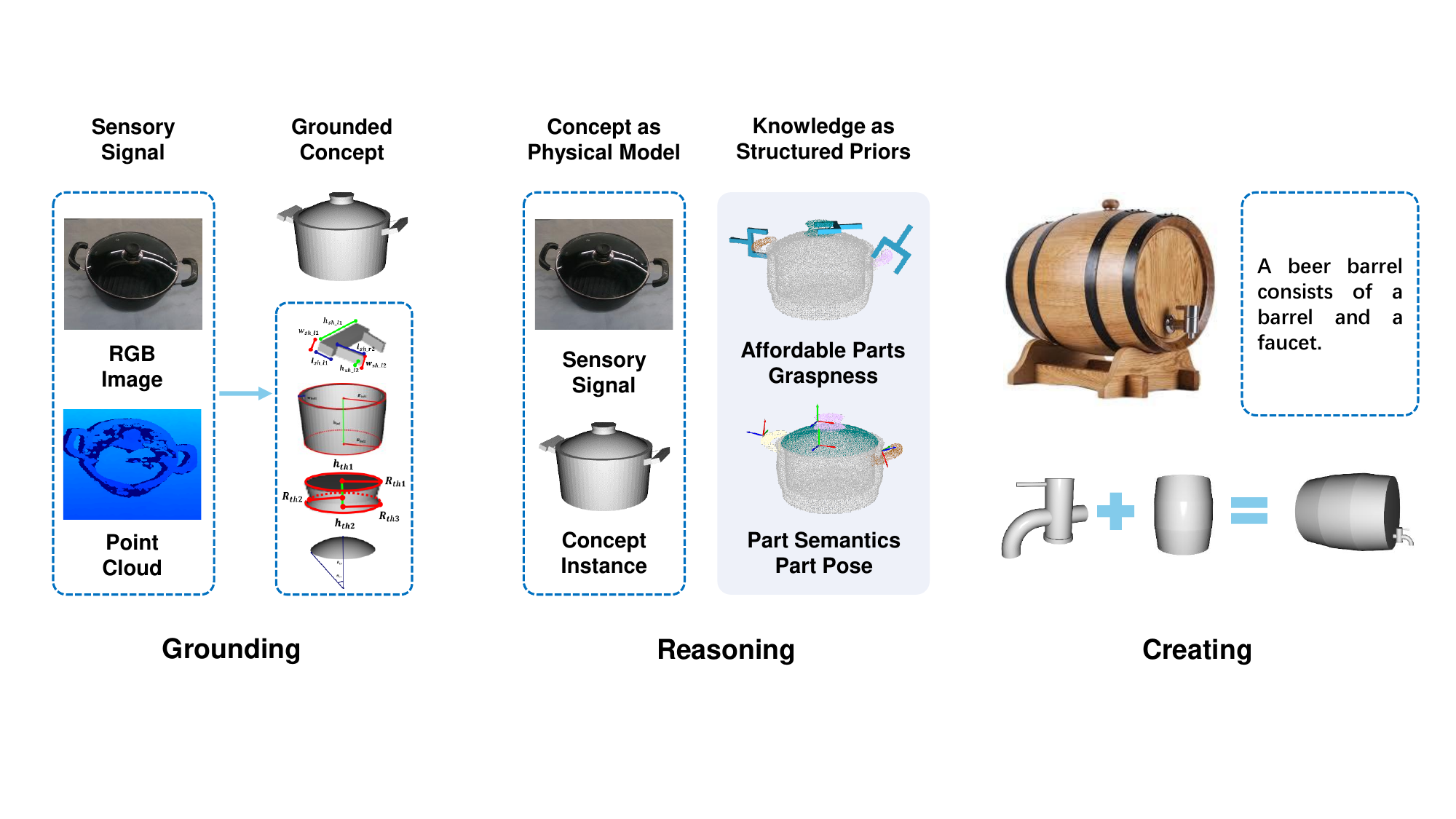}}
  \caption{Illustration of grounding, reasoning and creating with analytic concepts.}
  \label{fig:ForwardLearning}
\end{figure}

The forward process of concept learning aligns physical entities to analytic concepts, including grounding of analytic concepts, reasoning about the physical world, and creating non-existing concepts. A brief illustration is in Fig.~\ref{fig:ForwardLearning}.

\paragraph{Grounding} For a physical world signal, regardless of the modality of data used to express it (\textit{e.g.}, RGB, point clouds), a neural network can be trained to ground the analytical concepts with the physical world represented by this signal. The neural network typically comprises two main components, a concept parser to figure out what analytic concepts that best align with the physical world signal, and a concept parameter estimator to calculate the parameters of each analytic concept.

\paragraph{Reasoning} After grounding the analytical concepts with the physical world, the concept instances themselves can serve as a physical model for the target entity, while the knowledge within the concepts constitutes structured priors about the target. The physical model offers machine intelligence a fully observable state space of the physical target, and the structured priors depicts physical laws, environmental functioning principles, and commonsense human knowledge. Leveraging such information, machine intelligence can not only perform semantic reasoning (\textit{e.g.}, segmenting the refrigerator door) but also advance to numerical inference (\textit{e.g.}, determining whether a water bottle can stand upright in a refrigerator slot by considering the slot height and bottle dimensions).

\paragraph{Creating} There exists a finite cardinality of fundamental concepts in the physical domain, of which analytic concepts can typically be predefined by experts with small efforts. For more complex concepts composed of fundamental ones, we can certainly achieve their definition through expert specification, though more flexible approaches also exist. One approach involves fine-tuning large language models (LLMs) to leverage human-provided textual prompts describing complex concepts. These models can then utilize existing analytic concepts as a template library to generate novel complex ones. An alternative approach derives new analytic concepts by exploring and summarizing similar entities from data. 

\subsubsection{From Concepts to Physics (Inverse): Generating, Simulating and Planning}

The inverse process of concept learning conducts inference on analytical concepts and implements derived operations in the physical environment, including generating physical entities conditioned on analytic concepts, physical model simulating, and task planning.

\paragraph{Generating} Similar to the inverse process of grounding, we can generate physical entities by training generative models taking analytic concepts as conditions. Because analytic concepts serve as abstractions of physical world concepts, they can align with diverse data modalities captured by sensors. This enables the generation of data with a wide range of modalities such as RGB images, point clouds, 3D meshes, and also texts, while spanning from objects to entire scenes.

\paragraph{Simulating} The generating paradigm produces static data of physical entities. To reason about temporal physical world dynamics, one paradigm involves simulation through analytical concepts. Formally, a simulation process takes the sensory signal of the physical world, the current states of grounded analytic concepts and candidate action inputs, then models how the scenario evolves under the applied action. Such a process follows the formulation of world model while differs from some current techniques like Sora or Dreamer. Analytic concepts are fully physically grounded thus state transitions are calculated through mathematical methods that rigorously obey physical principles, systematically preventing physically implausible hallucinations that might arise from implicit neural networks.

\paragraph{Planning} The other paradigm to reason about temporal physical world dynamics is planning through analytical concepts. Different from simulating, planning does not take candidate action inputs. Instead, it takes a goal state as input and predicts a sequence of actions to achieve this goal. Also benefiting from the physically grounded property of analytic concepts, the planning process can be strictly guided by physical rules, avoiding physically implausible hallucinations in visual prediction.

\subsubsection{Close-loop Concept Learning (Bidirectional)}

Analytical concept's support for both forward and inverse learning allows us to implement bidirectional frameworks, achieving close-loop concept learning.

\paragraph{Active Concept Learning} For embodied agents like robots, a key advantage in physical world understanding lies in their capacity for interactive engagement with the physical environment. A type of closed-loop concept learning framework emerges through the agent's active interaction process. Machine intelligence grounds analytic concepts with physical world signals, using these concepts to guide embodied agents in physical interactions. In turn, these interactions changes the physical environment. By comparing observed physical changes with conceptual reasoning outcomes, machine intelligence generates feedback to iteratively refine the alignment accuracy between the conceptual space and the physical world. Put more simply, analytic concepts construct a concept-level world model of the physical world. As physical interactions deepen, the feedback gained from these interactions continuously refines the world model's consistency with the physical world.

\paragraph{Self-supervised Concept Learning} Another close-loop concept framework exists in the sequential nature of video data. For a video frame sequence, a forward concept learning network is first used to build a concept state representation of the current frame, and then the next states in the following frames are predicted according to an inverse concept learning network. By calculating the differences between the predicted states and the ground-truth ones in the video, the forward and inverse networks can be optimized. The entire process is self-supervised, capable of leveraging massive internet-scale video data for concept learning, thereby advancing scaling laws in physical world understanding.

\subsubsection{Data Preparation} 

To supervised train a concept learning neural network, it is necessary to collect a set of ``\textit{physical entity - analytic concept instance}" pairs. The sensory data of physical entity are often images and point clouds. In this part, we introduce two possible ways.

\paragraph{Manually Collection} The annotation of analytic concepts involves shape and pose parameters. These 3D labels are non-trivial for manually annotating through 2D interface. Therefore, we manually annotate 3D assets first, and then align the concept annotations with the images and point clouds of them captured in the physical world.

\paragraph{Procedural Generation} Another way is generating synthesized data through procedural generation, which is a technique of synthesizing data with generalized procedural rules. As analytic concepts are formally programmatic representations, they are naturally compatible with procedural generation. Specifically, we first write procedures based on analytic concepts to describe a series of objects and scenes, then randomly parameterize the procedures to synthesize concept instances, and finally resort to simulators or neural networks to render these concept instances into image or point cloud data. In this manner, the synthesized data inherently aligned with analytic concept annotations.

\section{Infrastructure of Analytic Concepts}

To facilitate the community and researchers to more easily implement algorithms using analytic concepts, we have developed a suite of infrastructure components centered around analytic concepts. Researchers can refer to \cite{sun2024conceptfactory} for our concept template library, concept annotation platform and ``\textit{Concept - Object}" dataset, refer to \cite{sun2024arti} for our object procedural generation approach, and \cite{sun2024discovering, sun2025physically} for baseline concept grounding approaches. 

\section{What's Next}

Up to this point, we have discussed about why we propose analytic concepts, how to develop and apply analytic concepts, and what we have established for analytic concepts. In the following version of this manuscript, we extend our discussion to the broad applications of analytic concepts and other more types of concept, and outline future research directions.

\bibliography{sn-bibliography}

\end{document}